\documentclass{bmvc2k}
\usepackage{paralist}
\usepackage{cleveref}
\usepackage{comment}
\usepackage{enumitem}
\usepackage{tikz}
\usepackage[super]{nth}
\usetikzlibrary{spy}
\setitemize{noitemsep,parsep=2pt,topsep=0pt}

\title{Multiview Image-based Hand Geometry Refinement using Differentiable Monte Carlo Ray Tracing}

\addauthor{Giorgos Karvounas}{gkarv@ics.forth.gr}{12}
\addauthor{Nikolaos Kyriazis}{kyriazis@ics.forth.gr}{1}
\addauthor{Iason Oikonomidis}{oikonom@ics.forth.gr}{1}
\addauthor{Aggeliki Tsoli}{aggeliki@ics.forth.gr}{1}
\addauthor{Antonis A. Argyros}{argyros@ics.forth.gr}{12}
\addinstitution{
Computational Vision and\\Robotics Laboratory\\
Institute of Computer Science\\
Foundation for Research and Technology - Hellas (FORTH)\\
Heraklion, Crete\\Greece
}
\addinstitution{
Computer Science Department\\
University of Crete\\
Heraklion, Crete\\Greece
}

\runninghead{KARVOUNAS, KYRIAZIS, OIKONOMIDIS, TSOLI, ARGYROS}{RAY-TRACING HAND}

\newcommand\hmm[1]{\ifnum\ifhmode\spacefactor\else2000\fi>1000 \uppercase{#1}\else#1\fi}
\newcommand{\raytracing}{\hmm{r}ay tracing}
\newcommand{\raytracer}{\hmm{r}ay tracer}
\newcommand{\raytracingfull}{\hmm{d}ifferentiable \raytracing}
\newcommand{\dataset}{InterHand2.6M}
\newcommand{\multiview}{\hmm{m}ultiview}

\def\eg{\emph{e.g}\bmvaOneDot}

\def\ie{\emph{i.e}\bmvaOneDot}

\def\etal{\emph{et al}\bmvaOneDot}

\makeatletter
\newcommand{\definetrim}[2]{%
  \define@key{Gin}{#1}[]{\setkeys{Gin}{trim=#2,clip}}%
}
\makeatother

\definetrim{exsff}{0 170 0 10}
\definetrim{twoC}{0 150 0 50}
\definetrim{rocker}{0 130 0 60}
\definetrim{fingers}{0 120 0 30}
\definetrim{fingers2}{0 150 0 0}
\definetrim{textures}{0 30 0 480}
\begin{document}

\maketitle


\begin{abstract}
{
The amount and quality of datasets and tools available in the research field of hand pose and shape estimation act as evidence to the significant progress that has been made. However, even the datasets of the highest quality, reported to date, have shortcomings in annotation. We propose a refinement approach, based on differentiable \raytracing{},  and demonstrate how a high-quality publicly available, multi-camera dataset of hands (\dataset{}) can become an even better dataset, with respect to annotation quality. \raytracingfull{} has not been employed so far to relevant problems and is hereby shown to be superior to the approximative alternatives that have been employed in the past. To tackle the lack of reliable ground truth, as far as quantitative evaluation is concerned, we resort to realistic synthetic data, to show that the improvement we induce is indeed significant. The same becomes evident in real data through visual evaluation.
}
\end{abstract}

\section{Introduction}
\label{sec:intro}

\begin{figure}
    \centering
    \includegraphics[width=\textwidth]{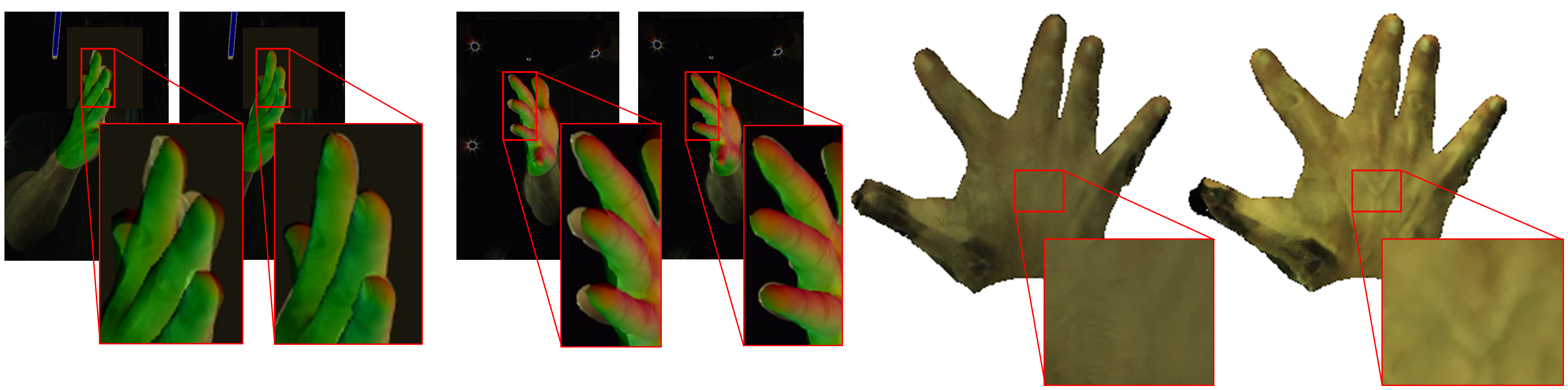}
    \vspace{-.4cm}
    \caption{Real datasets of hands, \eg \multiview{} dataset \dataset{} \cite{Moon_2020_ECCV_InterHand2.6M}, as well as results of state-of-the-art 3D hand pose/shape estimators, yield imperfect backprojections, which is evidence of error. We propose a method to improve the geometrical estimation of the hand, by exploiting color consistency, through a novel differentiable \raytracer{}, by Li \etal \cite{Li:2018:DMC}. In the three exemplar comparisons taken from the same instance, the left sides correspond to input state and the right sides to output state, after our method has been applied. As is evident, a better fitting geometry increases color consistency, which is reflected in the sharpness and definition of the appearance map, on the right.} 
    \label{fig:concept}
\end{figure}

Estimating the pose and shape of a human hand in 3D from RGB images is an important problem with numerous applications in human-computer interaction, augmented reality, robotics and more. State-of-the-art solutions have shown impressive results in real-world settings (see \Cref{sec:literature})\textcolor{red}{} with their remarkable performance heavily relying on advances in deep learning architectures \cite{simonyan2014very,he2016deep} 
and the availability of large datasets for training. Synthetic datasets have perfect ground truth information, but their lack of realism hinders generalization to the real domain. Real datasets model the real world directly, but come with noisy manual or automatically generated pose/shape annotations, that are nevertheless considered to be ground truth (see \Cref{fig:concept}). As a result, state-of-the-art 3D hand reconstructions from visual data exhibit mediocre image-to-model alignment. Increasing the amount of training data will improve the results only to the degree that the quality of the accompanied annotations is high. The ability to improve upon imperfect hand pose and shape estimates provided as image annotations or as the output of a neural network during training or testing can greatly benefit the performance of hand reconstruction methods.   

In this work, we propose a method for refining an imperfect estimate of the geometry of a hand using calibrated \multiview{} images as input. We follow a render-and-compare approach, and optimize the shape, pose and global orientation of the hand so that the discrepancy between the rendered and the observed hand, across multiple views, is minimized. The effectiveness of our method lies
\begin{inparaenum}[(a)]
  \item in the density of the information exploited as constraints to the problem, \ie\ all the pixels in all images that correspond to the hand, as well as
  \item the appropriateness of sampling in rendering, taken into account in the forward and the backward step of the optimization.
\end{inparaenum}

No previous work addresses unsupervised hand geometry refinement. However, the constancy of the appearance across images has been exploited in the context of hand or human pose estimation
in recent works~\cite{panteleris2017back,Moon_2020_ECCV_DeepHandMesh,bogo2014faust}.
Most rendering components invoked in the literature can be mapped to Pytorch3D~\cite{ravi2020pytorch3d}, which implements Neural Rendering~\cite{kato2018renderer} and Soft Rasterization~\cite{liu2020general}. Contrary to what has been employed so far, we make use of \raytracingfull{} \cite{Li:2018:DMC, nimier2019mitsuba}, and show that it delivers improved results, in comparison. For an impression of the gap between rasterization and \raytracing,{} and sampling correctness the reader is referred to \cite{glassner1989introduction}. We employ the differentiable Monte Carlo \raytracer{} by Li \etal \cite{Li:2018:DMC}, which provides a solution to gradient discontinuities at the edges. The latter practically implies that the benefits of \raytracing{} can also be exploited in the backward pass, instead of just the forward pass.


We showcase our method on a subset of the recently proposed real and large-scale dataset \dataset{}~\cite{Moon_2020_ECCV_InterHand2.6M}. \dataset{} contains imperfect ground truth annotations (see \Cref{fig:concept}) comprising 3D joint locations, derived in a semi-automatic way, and associated hand shape and pose descriptions, expressed via the MANO hand model, with a reported error of around $5 mm$. To circumvent the lack of ground truth data that are reliable enough for quantitative performance assessment, we resort to synthetic experimentation that closely resembles the setup in \dataset{}. We are able to successfully refine human hand estimates corresponding to a large variety of hand poses, shapes and appearances.

Our contributions can be summarized as follows.
\begin{itemize}
    \item We propose an effective unsupervised method for hand pose and shape refinement from \multiview{} images, given an initial estimate of the 3D hand geometry and, optionally, noisy target 3D joint locations. We complement the sparse noisy information about the 3D joint locations with rich image cues related to the appearance of the hand.
    \item We employ the proposed method to improve on a real dataset that is already on par with the state of the art with respect to annotation quality, while being richer in other aspects (number of samples, diversity in hand shape, pose, appearance, etc), yielding a dataset that is overall of even better quality.
    \item We employ, for the first time, a differentiable \raytracer{} to the problem of estimating the 3D pose and shape of hands, that is shown to be theoretically and practically superior to the rasterization methods that have been employed so far. 
    \item We provide a suite of synthetic experimentation to connect the degree of color consistency, as a function of geometry and appearance, with the improvement in 3D estimation of the hand, for the  described scenario.
\end{itemize}

\begin{figure}
    \begin{minipage}{1\textwidth}
    \includegraphics[clip, trim=0cm 11cm 0.5cm 11cm, width=1.00\textwidth]{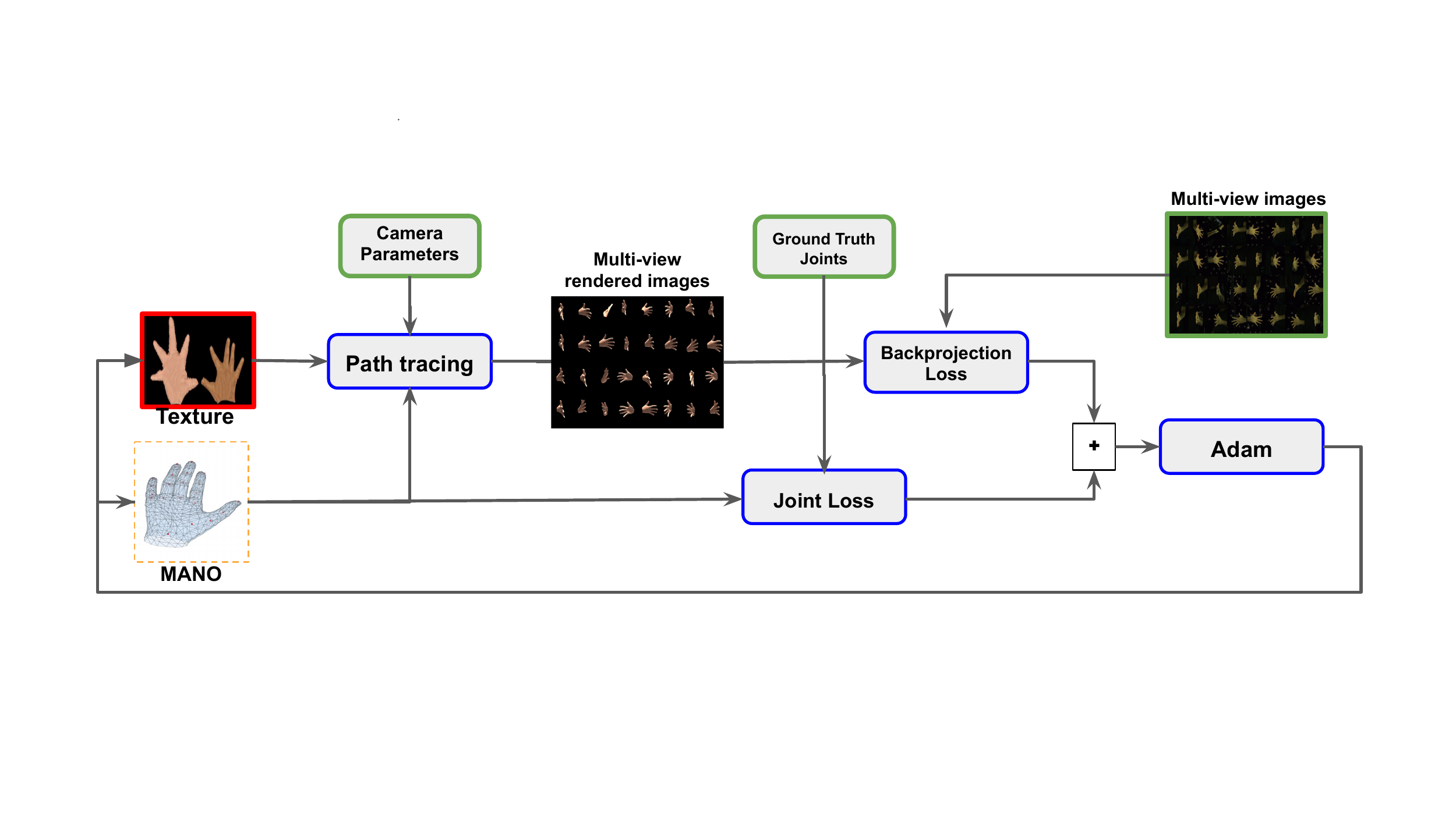}
    \vspace{3.5cm}
    \caption{To perform \multiview{} refinement of 3D hand geometry we compute the scalar difference between the backprojection of the currently estimated hand geometry and appearance and the observed input images. We update the estimates according to the gradient of the difference at the estimation point. Backprojection is performed using a \raytracer{}. 3D joint estimates, when available, can optionally be incorporated to aid search.
    }
    \label{fig:imperfect_data}
    \end{minipage}%
\end{figure}


\section{Literature Overview}
\label{sec:literature}

The task of 3D hand pose estimation from markerless visual input is a significant problem with numerous applications. It has been explored since the '90s~\cite{rehg1994digiteyes} and is still actively researched today~\cite{moon2018v2v, iqbal2018hand, han2020megatrack, spurr2021adversarial, spurr2021self, zimmermann2021contrastive, hampali2021handsformer}. Numerous important applications can be built based on hand pose estimation systems, including Human-Computer Interaction such as Augmented and Virtual Reality, Performance Capture and Assessment, and Sign Language Recognition and Translation.
The task is very challenging, due to several interdependent, complicating factors such as fast hand motions, self-occlusions, wide hand articulation range, and frequent interaction with the environment~\cite{erol2007vision}.

%
Given the difficulty of the problem, the early approaches predominantly relied on \multiview{} input to alleviate ambiguities arising from occlusions, an approach that has recently gained again popularity
~\cite{rehg1994digiteyes,de2008model,oikonomidis2010markerless,oikonomidis2011full,tzionas2016capturing,han2018online,han2020megatrack,smith2020constraining}. Stereo input was similarly explored~\cite{panteleris2017back,remilekun2017hand,li2019end}. With the wide availability of commodity depth sensors around 2010, the research also focused on monocular depth or RGBD input~\cite{oikonomidis2011efficient,keskin2013real,sridhar2013interactive,qian2014realtime,tang2014latent,makris2015model,tagliasacchi2015robust,oberweger2015training,wohlke2018model,moon2018v2v,ge2018robust}. However, shortly thereafter, the success of deep learning lead to the proliferation of robust systems that perform on monocular RGB input~\cite{romero2009monocular,zb2017hand,panteleris2018using,cai2018weakly,mueller2018ganerated,iqbal2018hand,ge20193d,gao2019variational,thompson2020hand,kulon2020weakly,cai20203d,zimmermann2021contrastive,hampali2021handsformer,spurr2021self,spurr2021adversarial}. More recently, event cameras have also been investigated~\cite{rudnev2020eventhands}.


The early approaches typically estimated the parameters of a kinematics skeleton with predefined bone lengths, either in the form of joint angles, or as 3D joint positions directly~\cite{rehg1994digiteyes,de2008model}.
The accuracy achieved with the availability of depth information yielded the first practical real-time systems~\cite{oikonomidis2011efficient}, and enabled on-the-fly adjustment of the bone lengths to the observed hand~\cite{makris2015model,tkach2017online}. The development and availability of a parametric hand pose and shape model~\cite{MANO:SIGGRAPHASIA:2017} in turn fueled research towards estimating the surface of the observed hand~\cite{wohlke2018model,ge20193d,boukhayma20193d,kulon2020weakly,Moon_2020_ECCV_DeepHandMesh}.
Finally, some recent works are presenting the first attempts to model the appearance of the hand to some extent~\cite{HTML_eccv2020, Moon_2020_ECCV_DeepHandMesh, chen2021model}. Their main focus is on the estimation of an appropriate hand model, while the affects of lighting and shadows are factored out.



Differentiable graphics renderers have lately gained traction as per their impact in many Computer Vision related tasks, too. Notably, such an approach was adopted specifically for hand pose estimation in the early work by De la Gorce \etal~\cite{de2008model}. In that work, a graphics pipeline was custom fitted with the ability to compute derivatives with respect to parameters of the scene. More recently, several generic renderers with built-in automatic differentiation have been presented~\cite{loper2014opendr,kato2018renderer,Li:2018:DMC, nimier2019mitsuba, loubet2019reparameterizing, liu2020general} with ~\cite{kato2018renderer} being the most widely adopted renderer. In this work, we propose the first method for inferring hand geometry from visual data that employs differentiable \raytracing{}~\cite{Li:2018:DMC}. This design choice allows for far more accurate modeling of the lighting in the scene than existing differentiable renderers, taking a big step towards handling the full complexity of real scenes. As we show in our experimental evaluation, the accuracy of estimating the hand geometry when we employ~\cite{Li:2018:DMC} instead of the state-of-the-art rasterization-based renderer~\cite{kato2018renderer} is significantly improved. Rendering via differentiable \raytracing{} can, additionally, benefit other domains that rely on analysis-by-synthesis approaches such as human pose estimation, 6D object pose estimation etc.

There is no previous work that focuses solely on unsupervised refinement of hand geometry. However, refinement of the hand geometry considering image evidence has been used within broader frameworks for reconstructing hands from visual data. For instance, a rendered hand is compared against the segmented silhouette of the observed hand ~\cite{baek2019pushing, boukhayma20193d} or its texture, where the hand texture is modeled using a uniform color prior~\cite{chen2021model} or a texture model~\cite{HTML_eccv2020}. Although using the appearance of the hand for matching carries more information than using the hand silhouette, inaccurate inference of the appearance of the hand in~\cite{chen2021model,HTML_eccv2020} led to no improvement on hand pose and shape estimation. We present an effective way for leveraging the appearance of the hand for pose and shape refinement via matching the appearance of the hand across multiple views. There exists limited previous work that has relied on photometric consistency among multiple views for hand pose estimation~\cite{Moon_2020_ECCV_DeepHandMesh,panteleris2017back} and human body estimation ~\cite{bogo2014faust,pavlakos2019texturepose}. The quality of the results is highly affected by the quality of the rendered hand or body geometries and the corresponding derivatives from the differentiable renderer. Note that dealing with human hands compared to the human body poses the extra challenge that the appearance of the fingers is self-similar.

\section{Method}
\label{sec:method}

The input to our method is a calibrated multiframe, $M$, an initial estimate of the hand geometry parameters, $h$, and optionally a set of 3D annotations, $J$, for the joints of the hand. $M$ consists of a set of RGB images $\{O_i\}, i=1,\ldots,N$ along with the extrinsic and intrinsic calibration information $\{C_i\}, i=1,\ldots,N$ of the corresponding cameras where $N$ is the number of cameras.
We represent the hand geometry using the MANO hand model~\cite{MANO:SIGGRAPHASIA:2017}, which is parameterized by $h = \left\{ {\beta ,\theta ,\tau ,\phi } \right\}$, where $\beta \in R^{10}$ is the identity shape, $\theta \in R^{45}$ the pose, $\tau \in R^3$ the 3D translation of the root and $\phi \in R^3$ the 3D rotation of the root. Any other differentiable representation of the hand that can be decoded into 3D geometry, \eg\cite{Moon_2020_ECCV_DeepHandMesh}, is employable, too. The optional 3D joint annotations, $J$, can be either a product of $h$ or a loosely coupled estimate, as is the case in the \dataset{} dataset. 

The output of our method is the finetuning, $\delta h$, of the hand geometry parameters, $h$, such that the new geometry derived from $h$ and $\delta h$, when backprojected, fits the observations $O_i$ more accurately than $h$. 
Similarly to previous work~\cite{panteleris2017back}, we rely on the color consistency assumption.
Contrary to~\cite{panteleris2017back}, we provide an alternative formulation of the problem, striving for higher fidelity.
More specifically, instead of requiring all rasterized pixels corresponding to projected geometry on each image to have the same color intensity in all reprojections on other images, we assume that each part of the geometry is consistently colored, through a common color map, \ie an image encoded in a set of variables $c$, and require that all backprojections, up to per camera $i$ color transforms $T_i$, match the observations. The main differences between the two approaches are that color consistency in our approach is a hard constraint, rather than a soft one, and that we do not compare rasterizations to color intensities interpolated in image space, which are prone to aliasing, but, rather, we require for each backprojected pixel, that constitutes an integration of samples, to match the observations in the corresponding pixels, which are also regarded as the integrals they really are. The integrals in reference regard the rendering equation \cite{kajiya1986rendering} which we compute through \cite{Li:2018:DMC}. This approach guarantees that the constraints drawn from observations are properly defined and their effects are properly backpropagated from image differentiation to geometrical changes, with sampling appropriateness being the highlight.

We, thus, define the following optimization problem:

\begin{equation}
\delta \hat h,\hat c, \hat T = \mathop {\arg \min }\limits_{\delta h,c,T} E\left( {\delta h,c,T} \right)
\; ,
\end{equation}
where
\begin{equation}
E\left( {\delta h,c,T} \right) = E_1(\delta h, c,T) + \lambda E_2(\delta h)
\; .
\label{eq:objective}
\end{equation}
%
%
The data term, $E_1$, of \Cref{eq:objective} amounts to the backprojection error between the renderings of the estimated geometry across multiple views and the corresponding observed \multiview{} images (Section~\ref{occ_aware}). The prior term, $E_2$, acts as a penalty for solutions that are far from the initial estimate of 3D joint locations (Section~\ref{joint_loss}).

\subsection{Data term}
\label{occ_aware}

We define the backprojection error as 

\begin{equation}
{E_1}\left( {\delta h,c,T} \right) = \sum\limits_{i = 1, \ldots ,N} {{f_1}\;\left( {{O_i},\;{T_i} \cdot R\left( {{G_v}\left( {h + \delta h} \right),V\left( c \right),{C_i}} \right)\;} \right)}
\; .
\label{eq:rec_err}
\end{equation}

\paragraph{Geometry:}
${G_v}\left( x \right)$ computes the 3D vertices of the geometry represented by $x$. Without loss of generality, in the presented work we implement this function through the MANO model \cite{Li:2018:DMC}. This means that $x \in {R^{61}}$, encoding global position and orientation, as well as non-dimensionality-reduced articulation.
Instead of our highly regularized approach, one might opt for a non-parametric representation, to also accommodate off-model deformations, without further changes in modeling.

\paragraph{Color mapping:}
Each surface element of the hand geometry is mapped to a pixel colour during rendering. This mapping takes the form of an image which is mapped to the geometry through a per pixel coordinate transform (see \cite{HTML_eccv2020}). We employ the same mechanism, but instead of a single image we consider a set $c = \{V_{low}, V_{high}\}$ of two images of different sizes: ${V_{low}} \in {R^{8 \times 8 \times 3}}$ and ${V_{high}} \in {R^{512 \times 512 \times 3}}$. Thus, $V\left( c \right) = {V_{low}} + {V_{high}}$. $V_{low}$ is scheduled early during optimization (see \Cref{sec:optimization}) and represents low frequency base color information. $V_{high}$ is added to the optimization schedule in a 2nd phase, as a detail residual on top of $V_{low}$, which through addition also includes medium and high frequency information. $V_{low}$ is scaled up in size so as to match $V_{high}$ prior to addition, through nearest neighbor interpolation.

\paragraph{Rendering:}
$R\left( {g,V,C} \right)$ is a function that renders an image, through \raytracing{}, out of the geometry $g$, using a camera specification  $C$, and a color map $V \in R^{H \times W \times 3}$ of height $H$ and width $W$, with known vertex associations to $g$. 
The renderer $R$, is implemented using the differentiable \raytracer{} by Li et al. \cite{Li:2018:DMC}. Color consistency is hardly met in real conditions. To account for slightly different colorization across cameras we allow the rendering result for camera $i$ to also vary with respect to a $3 \times 3$ linear color transform $T_i$, which is the same for all pixels corresponding to camera $i$. Best results have been attained for such transforms that are close but not equal to the identity transform.

\paragraph{Render-and-compare:}
The comparison between an observed image $o$ and a rendered hypothesis $r$ amounts to the aggregation of the per pixel color differences ($D_c$) and the differences between edge detections ($D_e$) computed on $o,r$:

\begin{equation}
    \begin{array}{*{20}{c}}
{{f_1}\left( {o,r} \right) = {D_c}\left( {o,r} \right) + {D_e}\left( {o,r} \right)}\\
{{D_c}\left( {o,r} \right) = \left| {o - r} \right|}\\
{{D_e}\left( {o,r} \right) = \left| {Canny\left( o \right)*K - Canny\left( r \right)*K} \right|}
\end{array}
\label{eq:dataterm}
\end{equation}

\noindent where $Canny\left( {.} \right)$ denotes the application of the Canny edge detector \cite{canny1986computational}, implemented through \cite{eriba2019kornia}. $K$ is a $11 \times 11$ Gaussian kernel of $\sigma=5$, that is convolved with the edge detections, prior to differentiation. The detected edges reflect the prominence of the hands against the background. The diffusive blurring counters the sharpness of the edge detection results and, thus, practically widens the basin of convergence.

\subsection{Prior term}
\label{joint_loss}

We define the joint error as 
\vspace{-.3cm}
\begin{equation}
\begin{array}{*{20}{c}}
{{E_2}\left( {\delta h} \right) = {f_2}\left( {J,\;{G_J}\left( {h + \delta h} \right)\;} \right),}&{{f_2}\left( {x,y} \right) = \left\{ {\begin{array}{*{20}{c}}
{0.5{{\left( {x - y} \right)}^2}/\gamma ,}&{\left| {x - y} \right| < \gamma }\\
{\left| {x - y} \right| - 0.5\gamma ,}&{{\rm{otherwise}}}
\end{array}} \right.}
\end{array},
\label{eq:joint_err}
\end{equation}
where ${G_J}\left( x \right)$ computes the 3D joint locations of the hand geometry $x$ which are compared against the target 3D joint locations, $J$. The error function $f_2$ is a coordinate-wise smooth L1 loss.
The purpose of this term is to favor solutions whose joint predictions are within $\gamma$ millimeters from $J$ in all axes. We use $\gamma=5$ in all our experiments.
It is worth noting that the weight of this term $\lambda$ can be zero for the majority of cases, but it is assigned a small value ($\lambda  = {10^{ - 3}}$) to robustify optimization against poses with high ambiguity (\eg closed fist) or cases where the initial starting point is significantly far from the sought optimum.
\subsection{Optimization}
\label{sec:optimization}

The optimization was carried out using the Adam algorithm \cite{kingma2014adam}. 
The simultaneous optimization of geometry and color is a very challenging problem, as the optimization dynamics yield a race between which variable set settles first, leading to unfavorable local optima. To remedy this, we introduce a warm-up phase of $50$ iterations where only variables $T$ (per-camera color transform) and $V_{low}$ (low detail color map) are considered. This warm-up phase is succeeded by an optimization of all the variables ($\delta h, c, T$), for a remainder of $150$ iterations. This scheduling has been experimentally shown to be robust for the case of the \dataset{} dataset. The implementation was done in PyTorch. For the average case of $30$ views in a multiframe, each optimization iteration requires $0.93s$ for the rasterization path and $13.77s$ for the \raytracing{} path with $16$ number of samples per pixel in both the forward and the backward pass, as estimated on a system powered by a \emph{NVidia Titan V GPU}. For the $200$ iterations we employed this yields a duration of $3.1min$ for the rasterization path and $45.9min$ for the \raytracing{} path to process a multiframe.

\section{Results}
\label{sec:res}

\subsection{Data and metrics}
\label{sec:metrics}

We evaluated our pose and shape refinement method on real data from the \dataset{} dataset~\cite{Moon_2020_ECCV_InterHand2.6M}. 
To circumvent the lack or reliable ground truth, as required for quantitative assessment, we additionally performed experiments on synthetic data with perfect ground truth.
These data closely simulated the \dataset{} dataset.
In the next sections, we report the following error metrics: 
\begin{enumerate}
\item \textbf{mean Euclidean per-vertex distance $\epsilon_v$} between the ground truth and predicted hand geometry (in $mm$), 
\item \textbf{backprojection error on image intensities $\epsilon_b$} ($D_c$ term in \Cref{eq:dataterm}) between the rendered projection of the refined geometry and the observed image, expressed in intensity levels (range $0-255$).   
\end{enumerate}

\subsection{Quantitative analysis on synthetic data}
\subsubsection{Geometry refinement from noisy input}

\begin{figure}[t]
    \centering
    \begin{minipage}{0.35\textwidth}
        \centering
        \includegraphics[width=\textwidth]{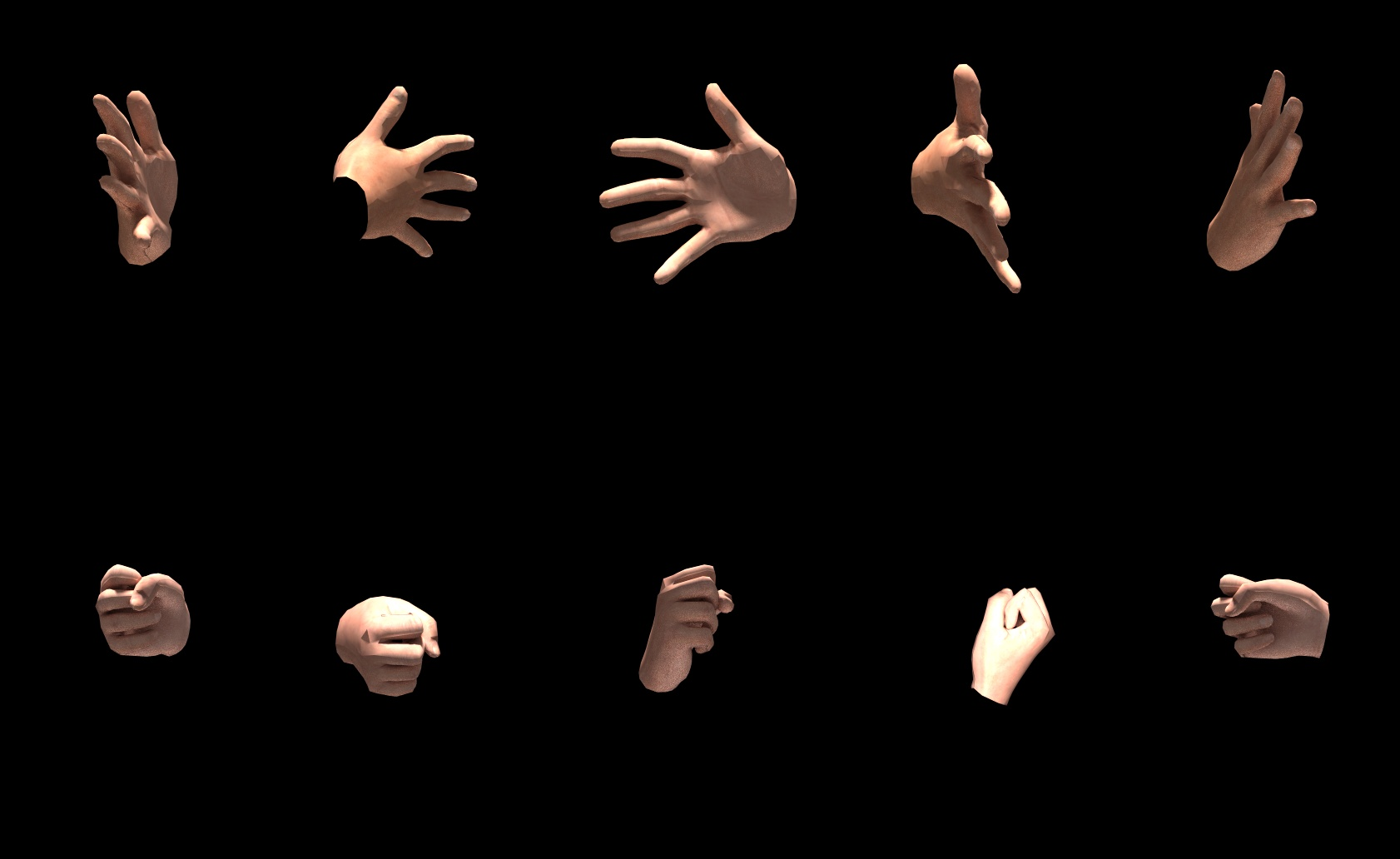}
    \end{minipage}
    \begin{minipage}{0.35\textwidth}
        \centering
        \includegraphics[width=\textwidth]{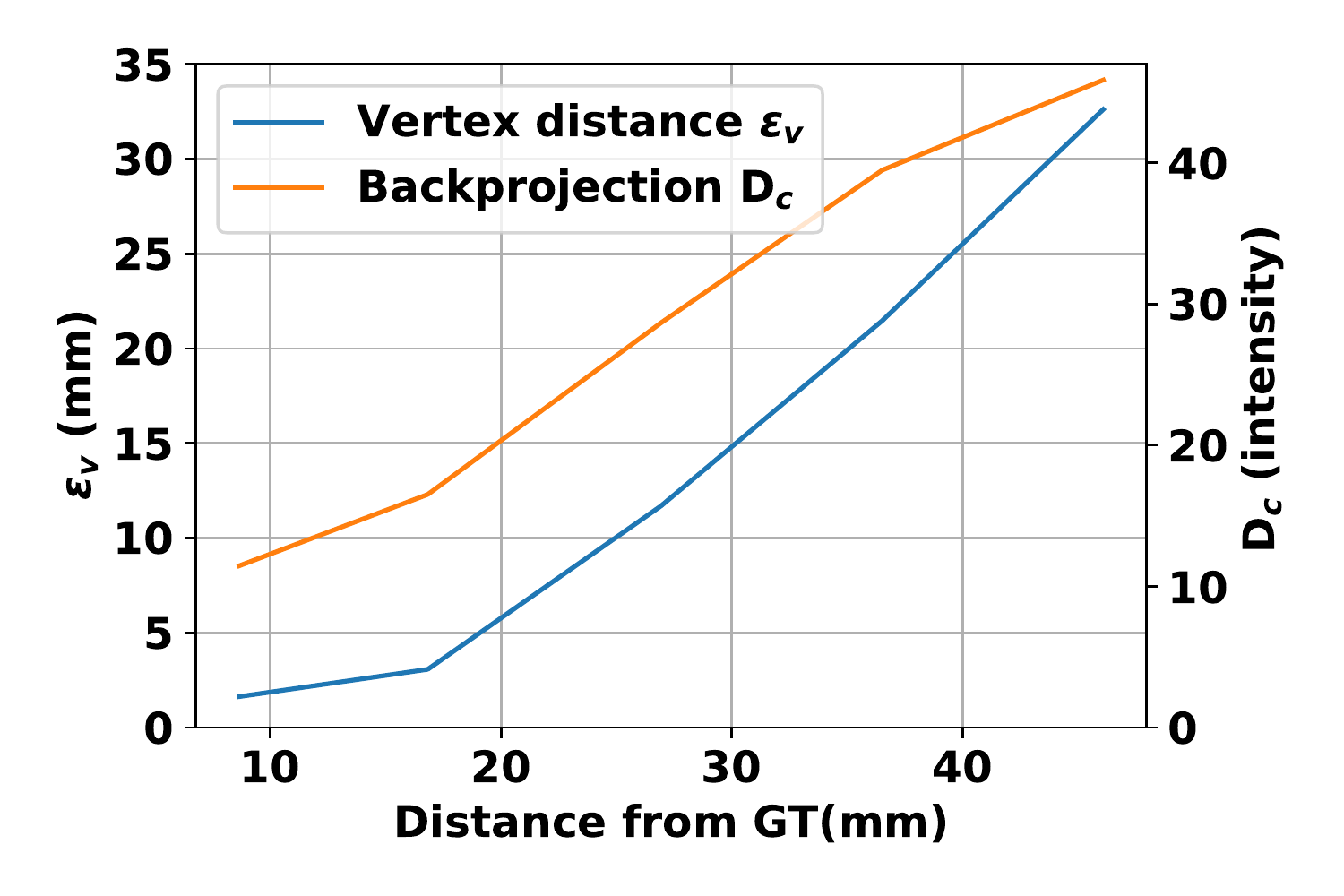}
    \end{minipage}
    \caption{\textbf{(left)} Examples of synthetic data. \textbf{(right)} Performance of our method on hand geometry refinement using synthetic data. The geometry of the hand is initialized at various distances from ground truth as shown in the $x$-axis. The $y$-axes show how the mean per-vertex error ($\epsilon_v$) and the backprojection error ($\epsilon_b$) after refinement exhibit similar behavior. Please refer to \Cref{sec:metrics} for the symbol definitions.
    }
    \label{fig:synthetic}
\end{figure}


We assessed the performance of our proposed method for varying levels of noise regarding the initial estimate of the hand geometry. We closely simulated the capture environment of the \dataset{} dataset, using the same number and location of virtual cameras (see \Cref{fig:synthetic}), and considered $5$ gestures, each performed by $2$ different subjects, resulting in a total of $10$ multiframes or $305$ independent frames (camera counts varied between $29$ and $32$). The gestures were chosen to depict varying degrees of self-occlusion. Noise was simulated by adding a random Gaussian vector to the MANO pose and shape parameters of the selected gestures and hand shapes. We defined the noise level as the mean per-vertex distance between the initial and perturbed hand geometry. The hand textures were generated using HTML~\cite{HTML_eccv2020} approximating the hand textures observed in \dataset{}. Examples of our synthetic data are shown in \Cref{fig:synthetic} on the left. 
On the right, we show the mean per-vertex error of the refined hand geometry and the corresponding backprojection error for all subjects and hand poses. The $x$ axis denotes the noise level. Note that we also considered noise levels beyond the average levels estimated in \dataset{} ($5mm$). We observe a monotonic relation between the two types of error, \ie the error on the estimated 3D geometry decreases as the backprojection error decreases. Similarly, we expect that decreasing the backprojection error on all views on real data will be accompanied by improvement in the actual 3D estimation.




\subsubsection{Ray tracing vs rasterization}
\label{sec:renderermatchup}

For \raytracing{} we use the differentiable Monte Carlo \raytracer{} by Li et al.~\cite{Li:2018:DMC}. As a rasterizer to compare with, we use the hard rasterization variant of Pytorch3D~\cite{ravi2020pytorch3d}, which is equivalent to Neural Rendering~\cite{kato2018renderer}, and has been commonly employed in the relevant literature, in its original form or in slight variations.

A few important notes are the following. The \raytracer{} by Li et al. \cite{Li:2018:DMC}, apart from implementing straightforward \raytracing{} and gradient computation, provides a continuous sampling solution for the edges, which in practice means that the basin of convergence for our backprojection problem is effectively widened. Other than that, the only hyperparameter of the employed \raytracer{} is the number of samples per pixel, which increases fidelity, \ie the approximation of the true integrals involved in the rendering equation. Hard rasterization does not have provision for discontinuities and widening of the convergence basin, but soft rasterization~\cite{liu2020general} does, by means of per pixel aggregation of neighboring information. The way soft rasterization is employed in optimization is that the ``softness'', controlled by a few parameters ($\gamma, \sigma$, blur amount, number of faces per pixel, etc.), is scheduled in a coarse-to-fine fashion, across iterations \cite{liu2020neural}. This means that at the last iterations, or when close to the sought solution, hard rasterization is effectively used. Since what is being discussed resides in this ballpark, we compare to hard rasterization.

\begin{figure}[t]
    \centering
    \begin{minipage}{0.3\textwidth}
        \centering
        \includegraphics[width=\textwidth]{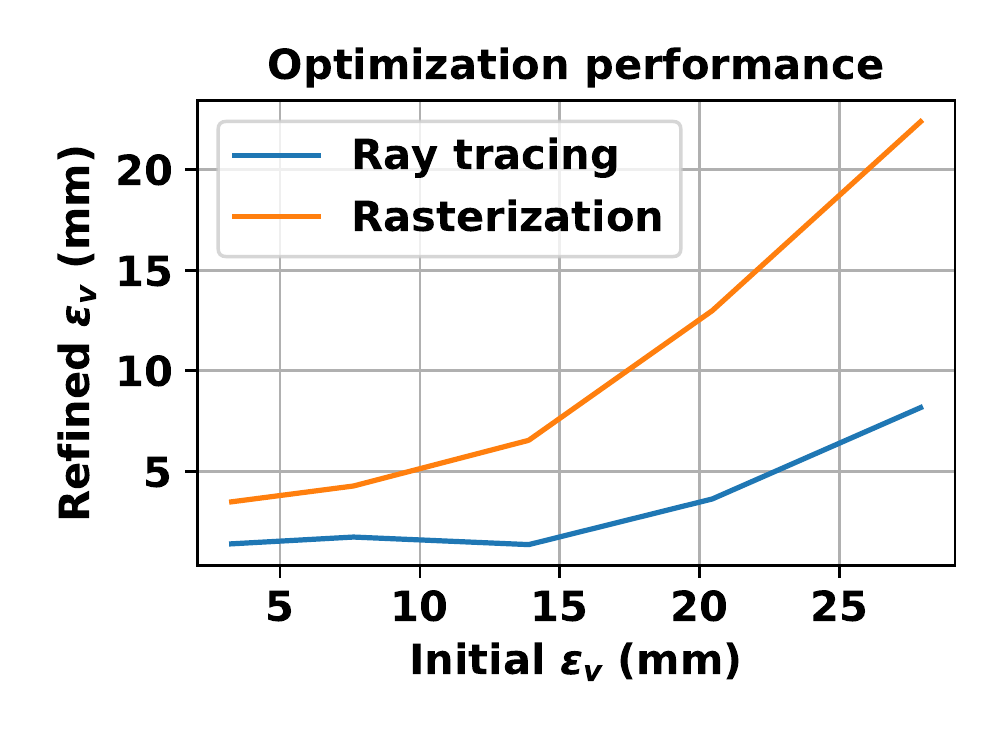}
    \end{minipage}
    \begin{minipage}{0.3\textwidth}
        \centering
        \includegraphics[width=\textwidth]{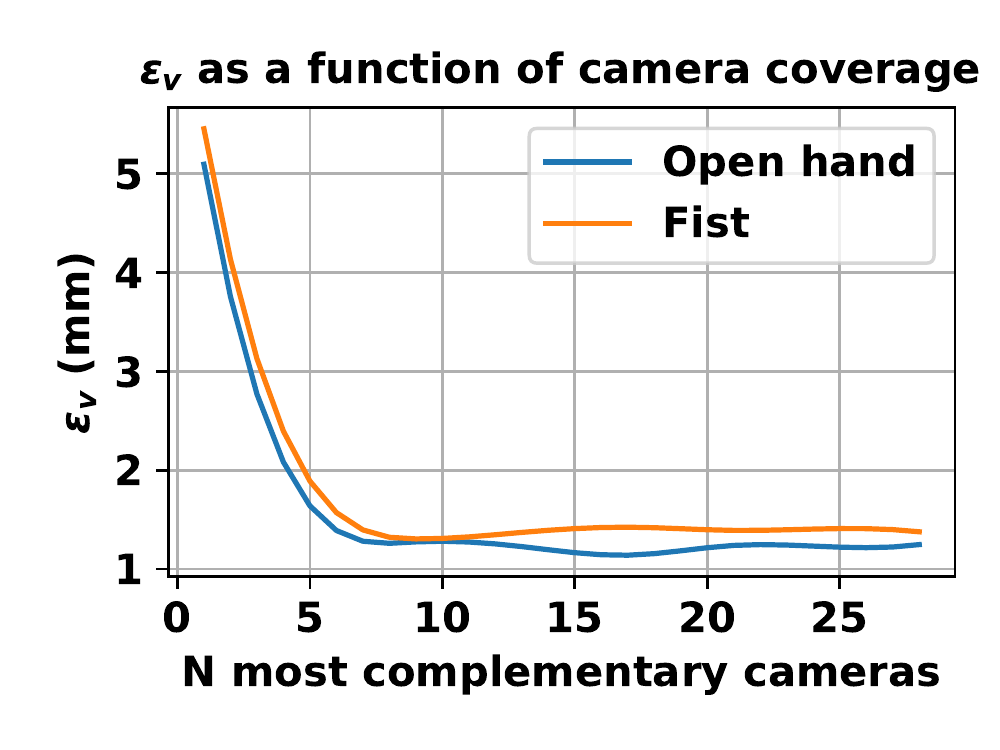}
    \end{minipage}
    \begin{minipage}{0.3\textwidth}
    
        \centering
        \includegraphics[width=\textwidth]{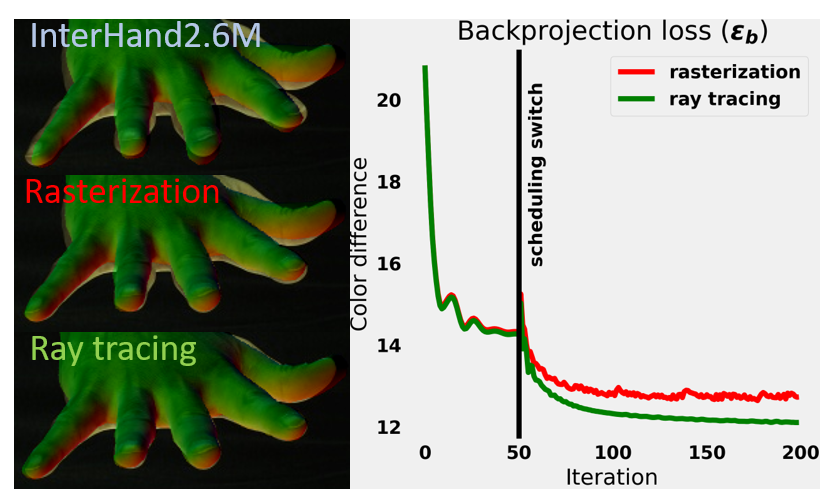}
    \end{minipage}
    \caption{\textbf{(left)} Comparison between \raytracing{} and rasterization in optimization. \textbf{(center)} $e_v$ as a function of optical coverage for two diverse scenarios. We observe that $7$ cameras are enough to get most of the benefit. \textbf{(right)} The superiority of \raytracing{} \cite{Li:2018:DMC} to rasterization \cite{ravi2020pytorch3d} in this task, all else being equal, is also evident in real data.}
    \label{fig:matchup}
\end{figure}

To compare between the two rendering methodologies we define the same, common problem, of refining a pose given calibrated multicamera RGB input. For the comparison everything remains the same, apart from the differentiable renderer itself. We define initial search locations that are increasingly further from the ground truth and try, through optimization, to recover the ground truth itself. The results shown in \Cref{fig:matchup} indicate that \raytracing{} is significantly more effective in the task described, with the advantage of simplicity and the disadvantage of computational cost, compared to rasterization.

\subsubsection{Camera coverage}
\label{sec:cameracoverage}

We evaluated our method by varying the number of camera views used as input. We noticed that the proposed method does not require all $32$ cameras in \dataset{} in order to incur a significant improvement on estimating the hand geometry. However, complementarity in coverage among the camera views was important, as expected. To demonstrate this we ordered the cameras in a new enumeration ${C_i},i = 1, \ldots ,N$, such that 
\vspace{-.1cm}
\begin{equation}
    comp\left( {{C_{1, \ldots ,i}}} \right) \leq comp\left( {{C_{1, \ldots ,i + j}}} \right),\forall i,j > 0,
\end{equation}
where $comp\left( {{C_{1, \ldots ,i}}} \right)$ computes the complementarity for cameras 1 to $i$, by aggregating estimates of how much of the surface of the hand becomes visible, on average. Put simply, we reordered the cameras such that every next camera incorporation maximizes the complementarity at every step. Given synthetic data as in \Cref{fig:matchup}, we initialized our optimization from poses that had $\epsilon_v=5 mm$. Then, we performed our optimization in runs, where each run added the next camera, in the order described earlier. Among all results we picked two diverse instances to summarize the effect of camera coverage. The first instance regards an open hand and the other regards a  fist. Both instances vary with respect to occlusions and ambiguity in observation. The results shown in \Cref{fig:matchup} indicate that, although more cameras are beneficial, most of the benefits can be gained by having $7$ complementary cameras.

\subsection{Qualitative analysis on real data}
\label{sec:qualitative}

\begin{figure}[!ht]

    \centering{
    \begin{tikzpicture}[spy using outlines={circle,red,magnification=3,size=0.7cm, connect spies}]
    \node {\includegraphics[fingers,clip,width=.14\textwidth]{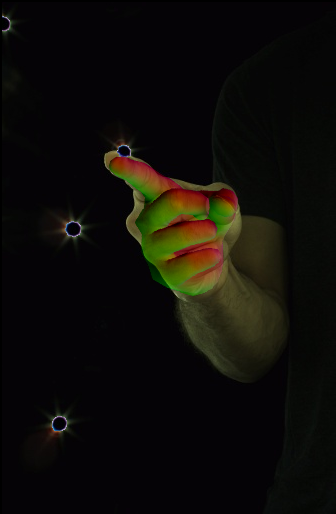}};
    \spy on (-.28,0.27) in node [left] at (-.3,1);
    \end{tikzpicture}
    \begin{tikzpicture}[spy using outlines={circle,red,magnification=3,size=0.7cm, connect spies}]
    \node {\includegraphics[fingers,clip,width=.14\textwidth]{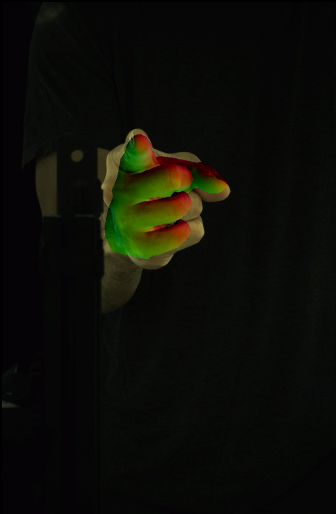}};
    \spy on (.25,0.15) in node [left] at (-.3,1.);
    \end{tikzpicture}
    \begin{tikzpicture}[spy using outlines={circle,red,magnification=3,size=0.7cm, connect spies}]
    \node {\includegraphics[fingers,clip,width=.14\textwidth]{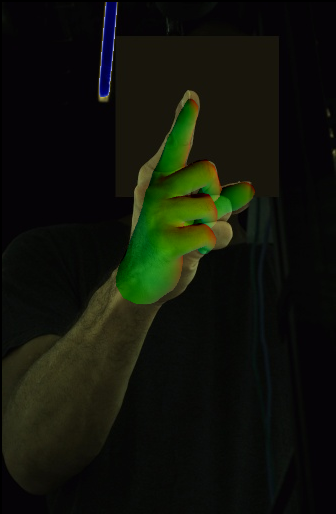}};
    \spy on (.25,0.15) in node [left] at (-.3,1.);
    \end{tikzpicture}
    \hspace{2px}
    \begin{tikzpicture}[spy using outlines={circle,red,magnification=3,size=0.7cm, connect spies}]
    \node {\includegraphics[fingers2,clip,width=.14\textwidth]{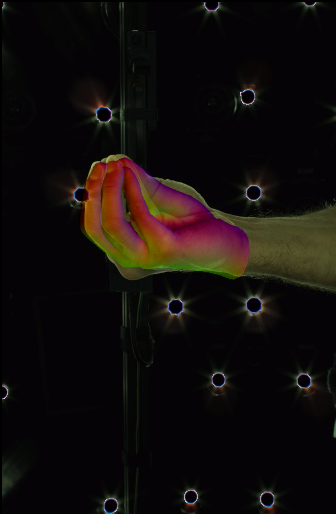}};
    \spy on (.1,-0.04) in node [left] at (-.3,1.);
    \end{tikzpicture}
    \begin{tikzpicture}[spy using outlines={circle,red,magnification=3,size=0.7cm, connect spies}]
    \node {\includegraphics[fingers,clip,width=.14\textwidth]{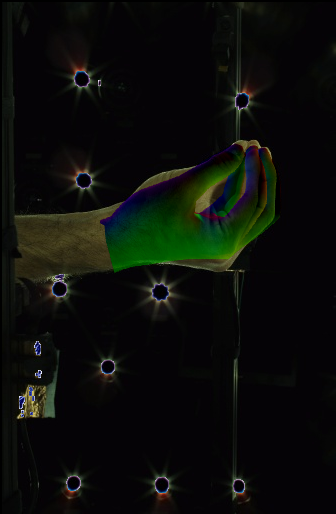}};
    \spy on (.45,0.36) in node [left] at (-.3,1.);
    \end{tikzpicture}
    \begin{tikzpicture}[spy using outlines={circle,red,magnification=3,size=0.7cm, connect spies}]
    \node {\includegraphics[fingers,clip,width=.14\textwidth]{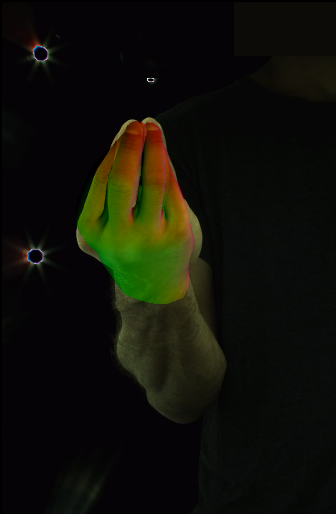}};
    \spy on (-.48,-0.18) in node [left] at (-.3,1.);
    \end{tikzpicture}

    \vspace{-0.5cm}
    
    \begin{tikzpicture}[spy using outlines={circle,green,magnification=3,size=0.7cm, connect spies}]
    \node {\includegraphics[fingers,clip,width=.14\textwidth]{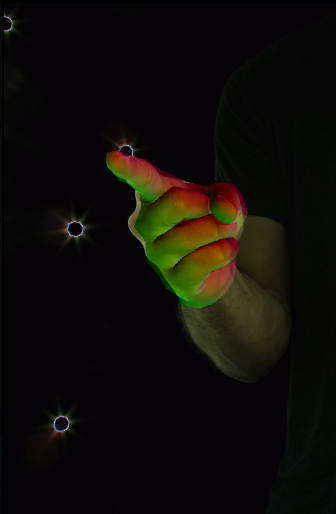}};
    \spy on (-.28,0.27) in node [left] at (-.3,1.);
    \end{tikzpicture}
    \begin{tikzpicture}[spy using outlines={circle,green,magnification=3,size=0.7cm, connect spies}]
    \node {\includegraphics[fingers,clip,width=.14\textwidth]{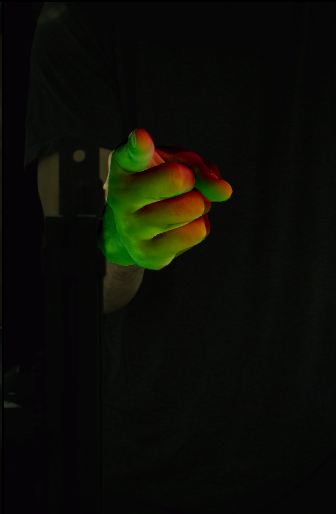}};
    \spy on (.25,0.15) in node [left] at (-.3,1.);
    \end{tikzpicture}
    \begin{tikzpicture}[spy using outlines={circle,green,magnification=3,size=0.7cm, connect spies}]
    \node {\includegraphics[fingers,clip,width=.14\textwidth]{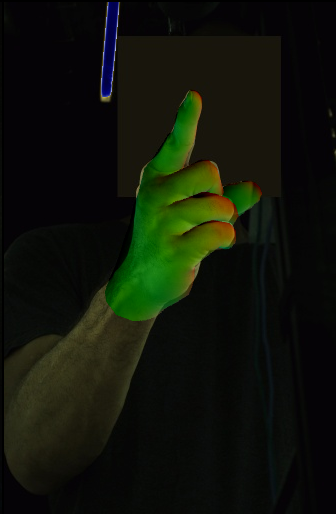}};
    \spy on (.25,0.15) in node [left] at (-.3,1.);
    \end{tikzpicture}
    \hspace{2px}
    \begin{tikzpicture}[spy using outlines={circle,green,magnification=3,size=0.7cm, connect spies}]
    \node {\includegraphics[fingers2,clip,width=.14\textwidth]{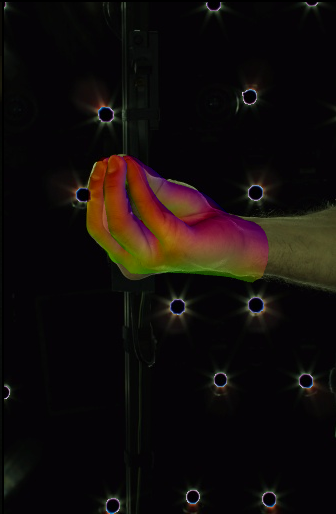}};
    \spy on (.1,-0.04) in node [left] at (-.3,1.);
    \end{tikzpicture}
    \begin{tikzpicture}[spy using outlines={circle,green,magnification=3,size=0.7cm, connect spies}]
    \node {\includegraphics[fingers,clip,width=.14\textwidth]{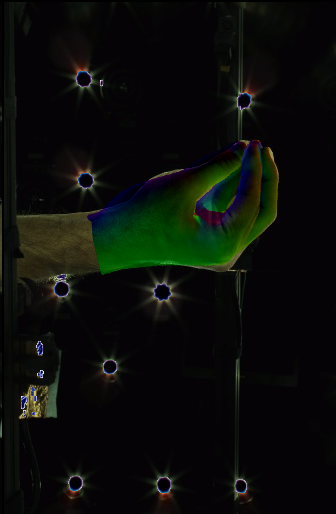}};
    \spy on (.45,0.36) in node [left] at (-.3,1.);
    \end{tikzpicture}
    \begin{tikzpicture}[spy using outlines={circle,green,magnification=3,size=0.7cm, connect spies}]
    \node {\includegraphics[fingers,clip,width=.14\textwidth]{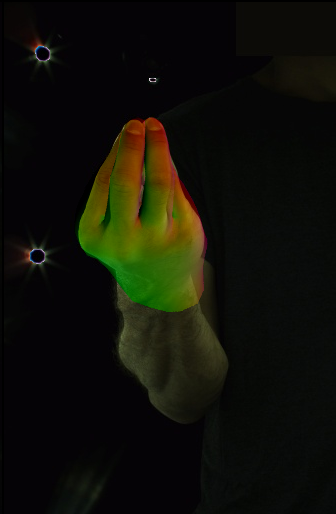}};
    \spy on (-.48,-0.18) in node [left] at (-.3,1.);
    \end{tikzpicture}

    \begin{tikzpicture}[spy using outlines={circle,red,magnification=3,size=0.7cm, connect spies}]
    \node {\includegraphics[fingers,clip,width=.14\textwidth]{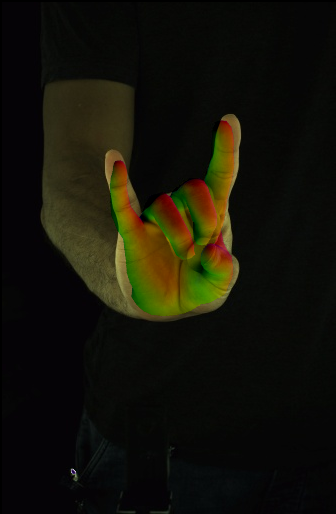}};
    \spy on (-.3,0.3) in node [left] at (-.3,1);
    \end{tikzpicture}
    \begin{tikzpicture}[spy using outlines={circle,red,magnification=3,size=0.7cm, connect spies}]
    \node {\includegraphics[fingers,clip,width=.14\textwidth]{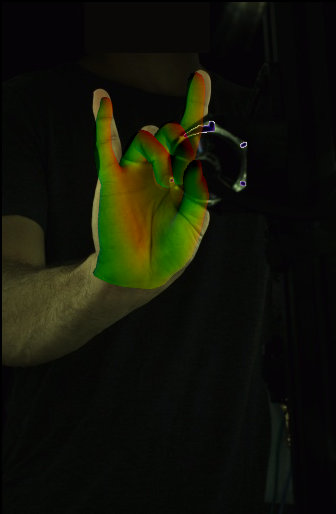}};
    \spy on (-0.39,0.1) in node [left] at (-.3,1.);
    \end{tikzpicture}
    \begin{tikzpicture}[spy using outlines={circle,red,magnification=3,size=0.7cm, connect spies}]
    \node {\includegraphics[fingers,clip,width=.14\textwidth]{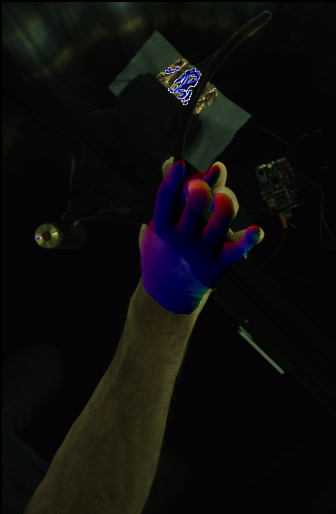}};
    \spy on (.3,0.2) in node [left] at (-.3,1.);
    \end{tikzpicture}
    \hspace{2px}
    \begin{tikzpicture}[spy using outlines={circle,red,magnification=3,size=0.7cm, connect spies}]
    \node {\includegraphics[fingers2,clip,width=.14\textwidth]{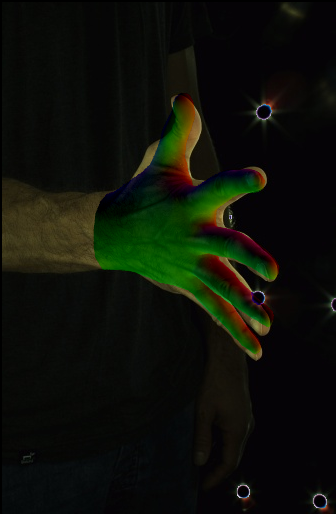}};
    \spy on (.3,-0.2) in node [left] at (-.3,1.);
    \end{tikzpicture}
    \begin{tikzpicture}[spy using outlines={circle,red,magnification=3,size=0.7cm, connect spies}]
    \node {\includegraphics[fingers,clip,width=.14\textwidth]{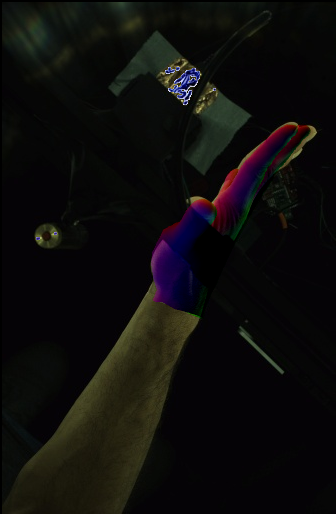}};
    \spy on (.45,0.36) in node [left] at (-.3,1.);
    \end{tikzpicture}
    \begin{tikzpicture}[spy using outlines={circle,red,magnification=3,size=0.7cm, connect spies}]
    \node {\includegraphics[fingers,clip,width=.14\textwidth]{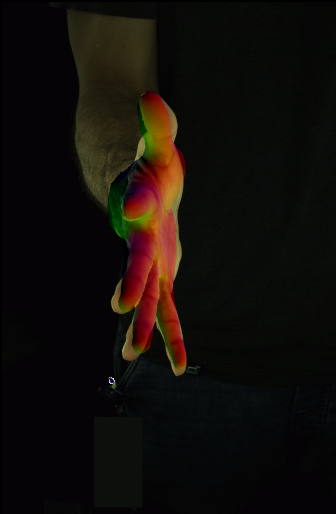}};
    \spy on (-.1,0.36) in node [left] at (-.3,1.);
    \end{tikzpicture}
    
    
    \vspace{-0.5cm}
    
    \begin{tikzpicture}[spy using outlines={circle,green,magnification=3,size=0.7cm, connect spies}]
    \node {\includegraphics[fingers,clip,width=.14\textwidth]{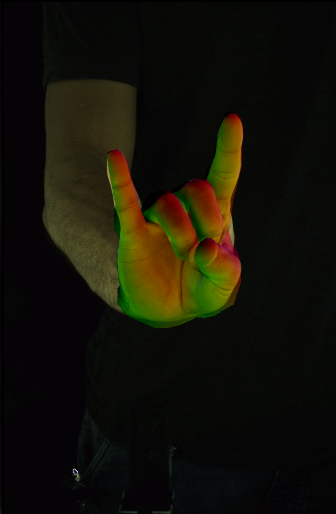}};
    \spy on (-.3,0.3) in node [left] at (-.3,1.);
    \end{tikzpicture}
    \begin{tikzpicture}[spy using outlines={circle,green,magnification=3,size=0.7cm, connect spies}]
    \node {\includegraphics[fingers,clip,width=.14\textwidth]{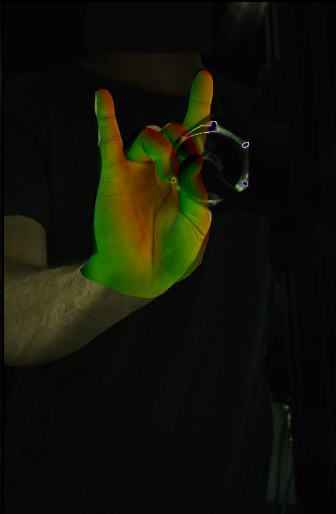}};
    \spy on (-0.39,0.1) in node [left] at (-.3,1.);
    \end{tikzpicture}
    \begin{tikzpicture}[spy using outlines={circle,green,magnification=3,size=0.7cm, connect spies}]
    \node {\includegraphics[fingers,clip,width=.14\textwidth]{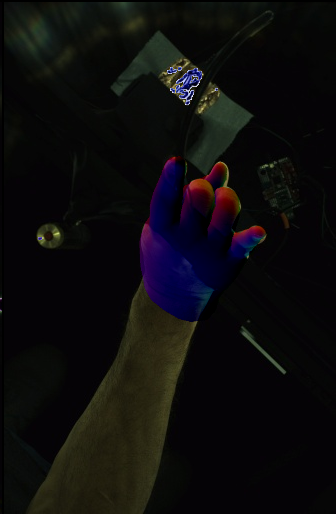}};
    \spy on (.3,0.2) in node [left] at (-.3,1.);
    \end{tikzpicture}
    \hspace{2px}
    \begin{tikzpicture}[spy using outlines={circle,green,magnification=3,size=0.7cm, connect spies}]
    \node {\includegraphics[fingers2,clip,width=.14\textwidth]{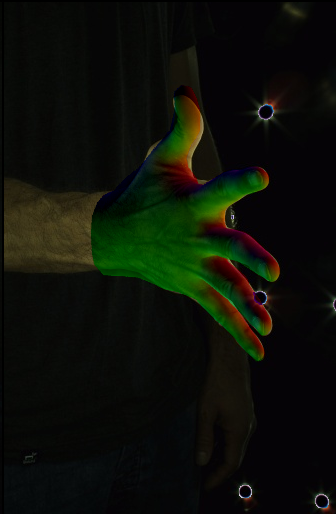}};
    \spy on (.3,-0.2) in node [left] at (-.3,1.);
    \end{tikzpicture}
    \begin{tikzpicture}[spy using outlines={circle,green,magnification=3,size=0.7cm, connect spies}]
    \node {\includegraphics[fingers,clip,width=.14\textwidth]{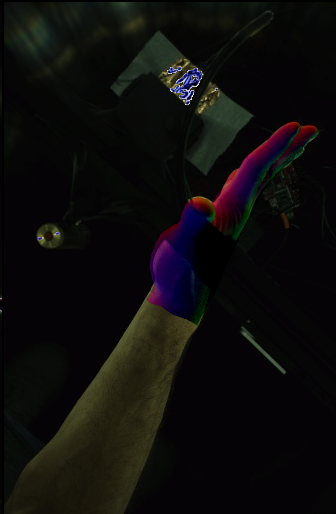}};
    \spy on (.45,0.36) in node [left] at (-.3,1.);
    \end{tikzpicture}
    \begin{tikzpicture}[spy using outlines={circle,green,magnification=3,size=0.7cm, connect spies}]
    \node {\includegraphics[fingers,clip,width=.14\textwidth]{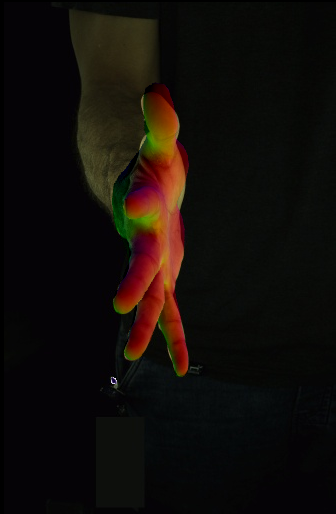}};
    \spy on (-.1,0.36) in node [left] at (-.3,1.);
    \end{tikzpicture}
    
    }
    
    \begin{flushleft}
    \vspace{-.4cm}
    \caption{Example results of our method using real data from \dataset{}. Odd rows show the hand annotations in \dataset{} based on MANO. Even rows show the refined hand geometry with our method.
    The proposed approach can handle a variety of hand poses and is robust to moderate hand occlusions.
    }
    \label{fig:backprojection}
    \end{flushleft}
\end{figure}

We evaluated our method on a subset of \dataset{}~\cite{Moon_2020_ECCV_InterHand2.6M}. \dataset{} is a large-scale dataset with real \multiview{} images depicting hands of various subjects performing sign language and everyday life gestures. The dataset provides annotations on 3D joint locations using a semi-automatic neural annotation method~\cite{moon2020neuralannot}. Given the 3D joint locations, pose and shape parameters of the MANO hand model were also extracted on the whole dataset with an average joint location error of $5mm$~\cite{Moon_2020_ECCV_InterHand2.6M}. However, when it comes to fine prediction of hand geometry, the provided annotations can mainly serve as references for the true location and geometry of the hand rather than ground truth. 
In this section, we provide qualitative evidence of our method improving the bundled annotations in \dataset{}. 
We ran our geometry refinement method on a subset of \dataset{} consisting of 7 subjects (6 male, 1 female) and a diverse set of poses, ranging from a highly self-occluded hand to a completely open hand. 
An illustration of the predicted hand geometry for various subjects and poses is shown in Figure~\ref{fig:backprojection}. The hand is colored based on the surface normals. It is worth noting that the comparative evaluation which validated our design choice of employing \cite{Li:2018:DMC} (see \Cref{sec:renderermatchup}) evidently generalizes to the case of real data, too (see \Cref{fig:matchup}). More qualitative results are provided in the supplementary material accompanying this paper.

\vspace{-0.2cm}

\vspace{-1.5cm}

\section{Discussion}
\label{sec:disc}

We proposed a method for \multiview{} refinement of the geometry of a hand, based on the assumption of color consistency among views and the employment of a powerful \raytracer{}. We presented a quantified analysis of the benefits of our method using a series of synthetic experiments and showed qualitatively that we improved upon the ground truth annotations of \dataset{}, a state-of-the-art dataset of real images with hands. What has been discussed so far has a straightforward impact on refining the output or the training procedure of supervised methods for hand geometry estimation. However, we find most exciting the fact that \raytracingfull, as employed here, for the first time, in the problem of hand pose/shape estimation, may open the door to self-supervised training and training in the wild. A differentiable \raytracer{} such as the one in \cite{Li:2018:DMC} presents the best chance at properly disentangling image formation, which is key to generalization, and, in turn,  self-supervision. The latter constitutes the basis of our future work in this direction.

\section*{Acknowledgements}
We gratefully acknowledge the support of NVIDIA Corporation with the donation of a Titan V GPU used for the execution of this research. The research project was supported by the Hellenic Foundation for Research and Innovation (H.F.R.I) under the "\nth{1} Call for H.F.R.I Research Projects to support Faculty members and Researchers and the procurement of high-cost research equipment" Project I.C.Humans, Number: 91.
\bibliography{bib}

\begin{thebibliography}{65}
\providecommand{\natexlab}[1]{#1}
\providecommand{\url}[1]{\texttt{#1}}
\expandafter\ifx\csname urlstyle\endcsname\relax
  \providecommand{\doi}[1]{doi: #1}\else
  \providecommand{\doi}{doi: \begingroup \urlstyle{rm}\Url}\fi

\bibitem[Baek et~al.(2019)Baek, Kim, and Kim]{baek2019pushing}
Seungryul Baek, Kwang~In Kim, and Tae-Kyun Kim.
\newblock Pushing the envelope for rgb-based dense 3d hand pose estimation via
  neural rendering.
\newblock In \emph{Proceedings of the IEEE/CVF Conference on Computer Vision
  and Pattern Recognition}, pages 1067--1076, 2019.

\bibitem[Bogo et~al.(2014)Bogo, Romero, Loper, and Black]{bogo2014faust}
Federica Bogo, Javier Romero, Matthew Loper, and Michael~J Black.
\newblock Faust: Dataset and evaluation for 3d mesh registration.
\newblock In \emph{Proceedings of the IEEE Conference on Computer Vision and
  Pattern Recognition}, pages 3794--3801, 2014.

\bibitem[Boukhayma et~al.(2019)Boukhayma, Bem, and Torr]{boukhayma20193d}
Adnane Boukhayma, Rodrigo~de Bem, and Philip~HS Torr.
\newblock 3d hand shape and pose from images in the wild.
\newblock In \emph{Proceedings of the IEEE/CVF Conference on Computer Vision
  and Pattern Recognition}, pages 10843--10852, 2019.

\bibitem[Cai et~al.(2018)Cai, Ge, Cai, and Yuan]{cai2018weakly}
Yujun Cai, Liuhao Ge, Jianfei Cai, and Junsong Yuan.
\newblock Weakly-supervised 3d hand pose estimation from monocular rgb images.
\newblock In \emph{Proceedings of the European Conference on Computer Vision
  (ECCV)}, pages 666--682, 2018.

\bibitem[Cai et~al.(2020)Cai, Ge, Cai, Magnenat-Thalmann, and Yuan]{cai20203d}
Yujun Cai, Liuhao Ge, Jianfei Cai, Nadia Magnenat-Thalmann, and Junsong Yuan.
\newblock 3d hand pose estimation using synthetic data and weakly labeled rgb
  images.
\newblock \emph{IEEE transactions on pattern analysis and machine
  intelligence}, 2020.

\bibitem[Canny(1986)]{canny1986computational}
John Canny.
\newblock A computational approach to edge detection.
\newblock \emph{IEEE Transactions on pattern analysis and machine
  intelligence}, pages 679--698, 1986.

\bibitem[Chen et~al.(2021)Chen, Tu, Kang, Bao, Zhang, Zhe, Chen, and
  Yuan]{chen2021model}
Yujin Chen, Zhigang Tu, Di~Kang, Linchao Bao, Ying Zhang, Xuefei Zhe, Ruizhi
  Chen, and Junsong Yuan.
\newblock Model-based 3d hand reconstruction via self-supervised learning.
\newblock In \emph{Proceedings of the IEEE/CVF Conference on Computer Vision
  and Pattern Recognition}, pages 10451--10460, 2021.

\bibitem[de~La~Gorce et~al.(2008)de~La~Gorce, Paragios, and Fleet]{de2008model}
Martin de~La~Gorce, Nikos Paragios, and David~J Fleet.
\newblock Model-based hand tracking with texture, shading and self-occlusions.
\newblock In \emph{2008 IEEE Conference on Computer Vision and Pattern
  Recognition}, 2008.

\bibitem[Erol et~al.(2007)Erol, Bebis, Nicolescu, Boyle, and
  Twombly]{erol2007vision}
Ali Erol, George Bebis, Mircea Nicolescu, Richard~D Boyle, and Xander Twombly.
\newblock Vision-based hand pose estimation: A review.
\newblock \emph{Computer Vision and Image Understanding}, 108\penalty0
  (1-2):\penalty0 52--73, 2007.

\bibitem[Gao et~al.(2019)Gao, Wang, Falco, Navab, and
  Tombari]{gao2019variational}
Yafei Gao, Yida Wang, Pietro Falco, Nassir Navab, and Federico Tombari.
\newblock Variational object-aware 3-d hand pose from a single rgb image.
\newblock \emph{IEEE Robotics and Automation Letters}, 4\penalty0 (4):\penalty0
  4239--4246, 2019.

\bibitem[Ge et~al.(2018)Ge, Liang, Yuan, and Thalmann]{ge2018robust}
Liuhao Ge, Hui Liang, Junsong Yuan, and Daniel Thalmann.
\newblock Robust 3d hand pose estimation from single depth images using
  multi-view cnns.
\newblock \emph{IEEE Transactions on Image Processing}, 27\penalty0
  (9):\penalty0 4422--4436, 2018.

\bibitem[Ge et~al.(2019)Ge, Ren, Li, Xue, Wang, Cai, and Yuan]{ge20193d}
Liuhao Ge, Zhou Ren, Yuncheng Li, Zehao Xue, Yingying Wang, Jianfei Cai, and
  Junsong Yuan.
\newblock 3d hand shape and pose estimation from a single rgb image.
\newblock In \emph{Proceedings of the IEEE/CVF Conference on Computer Vision
  and Pattern Recognition}, pages 10833--10842, 2019.

\bibitem[Glassner(1989)]{glassner1989introduction}
Andrew~S Glassner.
\newblock \emph{An introduction to ray tracing}.
\newblock Morgan Kaufmann, 1989.

\bibitem[Hampali et~al.(2021)Hampali, Sarkar, Rad, and
  Lepetit]{hampali2021handsformer}
Shreyas Hampali, Sayan~Deb Sarkar, Mahdi Rad, and Vincent Lepetit.
\newblock Handsformer: Keypoint transformer for monocular 3d pose estimation of
  hands and object in interaction.
\newblock \emph{arXiv preprint arXiv:2104.14639}, 2021.

\bibitem[Han et~al.(2018)Han, Liu, Wang, Ye, Twigg, and Kin]{han2018online}
Shangchen Han, Beibei Liu, Robert Wang, Yuting Ye, Christopher~D Twigg, and
  Kenrick Kin.
\newblock Online optical marker-based hand tracking with deep labels.
\newblock \emph{ACM Transactions on Graphics (TOG)}, 2018.

\bibitem[Han et~al.(2020)Han, Liu, Cabezas, Twigg, Zhang, Petkau, Yu, Tai,
  Akbay, Wang, et~al.]{han2020megatrack}
Shangchen Han, Beibei Liu, Randi Cabezas, Christopher~D Twigg, Peizhao Zhang,
  Jeff Petkau, Tsz-Ho Yu, Chun-Jung Tai, Muzaffer Akbay, Zheng Wang, et~al.
\newblock Megatrack: monochrome egocentric articulated hand-tracking for
  virtual reality.
\newblock \emph{ACM Transactions on Graphics (TOG)}, 2020.

\bibitem[He et~al.(2016)He, Zhang, Ren, and Sun]{he2016deep}
Kaiming He, Xiangyu Zhang, Shaoqing Ren, and Jian Sun.
\newblock Deep residual learning for image recognition.
\newblock In \emph{Proceedings of the IEEE conference on computer vision and
  pattern recognition}, pages 770--778, 2016.

\bibitem[Iqbal et~al.(2018)Iqbal, Molchanov, Gall, and Kautz]{iqbal2018hand}
Umar Iqbal, Pavlo Molchanov, Thomas Breuel~Juergen Gall, and Jan Kautz.
\newblock Hand pose estimation via latent 2.5d heatmap regression.
\newblock In \emph{Proceedings of the European Conference on Computer Vision
  (ECCV)}, pages 118--134, 2018.

\bibitem[Kajiya(1986)]{kajiya1986rendering}
James~T Kajiya.
\newblock The rendering equation.
\newblock In \emph{Proceedings of the 13th annual conference on Computer
  graphics and interactive techniques}, pages 143--150, 1986.

\bibitem[Kato et~al.(2018)Kato, Ushiku, and Harada]{kato2018renderer}
Hiroharu Kato, Yoshitaka Ushiku, and Tatsuya Harada.
\newblock Neural 3d mesh renderer.
\newblock \emph{The IEEE Conference on Computer Vision and Pattern Recognition
  (CVPR)}, 2018.

\bibitem[Keskin et~al.(2013)Keskin, K{\i}ra{\c{c}}, Kara, and
  Akarun]{keskin2013real}
Cem Keskin, Furkan K{\i}ra{\c{c}}, Yunus~Emre Kara, and Lale Akarun.
\newblock Real time hand pose estimation using depth sensors.
\newblock In \emph{Consumer depth cameras for computer vision}, pages 119--137.
  Springer, 2013.

\bibitem[Kingma and Ba(2014)]{kingma2014adam}
Diederik~P Kingma and Jimmy Ba.
\newblock Adam: A method for stochastic optimization.
\newblock \emph{arXiv preprint arXiv:1412.6980}, 2014.

\bibitem[Kulon et~al.(2020)Kulon, Guler, Kokkinos, Bronstein, and
  Zafeiriou]{kulon2020weakly}
Dominik Kulon, Riza~Alp Guler, Iasonas Kokkinos, Michael~M Bronstein, and
  Stefanos Zafeiriou.
\newblock Weakly-supervised mesh-convolutional hand reconstruction in the wild.
\newblock In \emph{Proceedings of the IEEE/CVF Conference on Computer Vision
  and Pattern Recognition}, pages 4990--5000, 2020.

\bibitem[Li et~al.(2018)Li, Aittala, Durand, and Lehtinen]{Li:2018:DMC}
Tzu-Mao Li, Miika Aittala, Fr{\'e}do Durand, and Jaakko Lehtinen.
\newblock Differentiable monte carlo ray tracing through edge sampling.
\newblock \emph{ACM Trans. Graph. (Proc. SIGGRAPH Asia)}, 2018.

\bibitem[Li et~al.(2019)Li, Xue, Wang, Ge, Ren, and Rodriguez]{li2019end}
Yuncheng Li, Zehao Xue, Yingying Wang, Liuhao Ge, Zhou Ren, and Jonathan
  Rodriguez.
\newblock End-to-end 3d hand pose estimation from stereo cameras.
\newblock In \emph{BMVC}, volume~1, page~2, 2019.

\bibitem[Liu et~al.(2020{\natexlab{a}})Liu, Xu, Habermann, Zollhoefer, Bernard,
  Kim, Wang, and Theobalt]{liu2020neural}
Lingjie Liu, Weipeng Xu, Marc Habermann, Michael Zollhoefer, Florian Bernard,
  Hyeongwoo Kim, Wenping Wang, and Christian Theobalt.
\newblock Neural human video rendering by learning dynamic textures and
  rendering-to-video translation.
\newblock \emph{IEEE Transactions on Visualization and Computer Graphics},
  2020{\natexlab{a}}.

\bibitem[Liu et~al.(2020{\natexlab{b}})Liu, Li, Chen, and Li]{liu2020general}
Shichen Liu, Tianye Li, Weikai Chen, and Hao Li.
\newblock A general differentiable mesh renderer for image-based 3d reasoning.
\newblock \emph{IEEE Transactions on Pattern Analysis and Machine
  Intelligence}, 2020{\natexlab{b}}.

\bibitem[Loper and Black(2014)]{loper2014opendr}
Matthew~M Loper and Michael~J Black.
\newblock Opendr: An approximate differentiable renderer.
\newblock In \emph{European Conference on Computer Vision}, pages 154--169.
  Springer, 2014.

\bibitem[Loubet et~al.(2019)Loubet, Holzschuch, and
  Jakob]{loubet2019reparameterizing}
Guillaume Loubet, Nicolas Holzschuch, and Wenzel Jakob.
\newblock Reparameterizing discontinuous integrands for differentiable
  rendering.
\newblock \emph{ACM Transactions on Graphics (TOG)}, 38\penalty0 (6):\penalty0
  1--14, 2019.

\bibitem[Makris and Argyros(2015)]{makris2015model}
Alexandros Makris and A~Argyros.
\newblock Model-based 3d hand tracking with on-line hand shape adaptation.
\newblock In \emph{Proc. BMVC}, pages 77--1, 2015.

\bibitem[Moon and Lee(2020)]{moon2020neuralannot}
Gyeongsik Moon and Kyoung~Mu Lee.
\newblock Neuralannot: Neural annotator for in-the-wild expressive 3d human
  pose and mesh training sets.
\newblock \emph{arXiv preprint arXiv:2011.11232}, 2020.

\bibitem[Moon et~al.(2018)Moon, Chang, and Lee]{moon2018v2v}
Gyeongsik Moon, Ju~Yong Chang, and Kyoung~Mu Lee.
\newblock V2v-posenet: Voxel-to-voxel prediction network for accurate 3d hand
  and human pose estimation from a single depth map.
\newblock In \emph{Proceedings of the IEEE conference on computer vision and
  pattern Recognition}, pages 5079--5088, 2018.

\bibitem[Moon et~al.(2020{\natexlab{a}})Moon, Shiratori, and
  Lee]{Moon_2020_ECCV_DeepHandMesh}
Gyeongsik Moon, Takaaki Shiratori, and Kyoung~Mu Lee.
\newblock Deephandmesh: A weakly-supervised deep encoder-decoder framework for
  high-fidelity hand mesh modeling.
\newblock In \emph{European Conference on Computer Vision (ECCV)},
  2020{\natexlab{a}}.

\bibitem[Moon et~al.(2020{\natexlab{b}})Moon, Yu, Wen, Shiratori, and
  Lee]{Moon_2020_ECCV_InterHand2.6M}
Gyeongsik Moon, Shoou-I Yu, He~Wen, Takaaki Shiratori, and Kyoung~Mu Lee.
\newblock Interhand2.6m: A dataset and baseline for 3d interacting hand pose
  estimation from a single rgb image.
\newblock In \emph{European Conference on Computer Vision (ECCV)},
  2020{\natexlab{b}}.

\bibitem[Mueller et~al.(2018)Mueller, Bernard, Sotnychenko, Mehta, Sridhar,
  Casas, and Theobalt]{mueller2018ganerated}
Franziska Mueller, Florian Bernard, Oleksandr Sotnychenko, Dushyant Mehta,
  Srinath Sridhar, Dan Casas, and Christian Theobalt.
\newblock Ganerated hands for real-time 3d hand tracking from monocular rgb.
\newblock In \emph{Proceedings of the IEEE Conference on Computer Vision and
  Pattern Recognition}, pages 49--59, 2018.

\bibitem[Nimier-David et~al.(2019)Nimier-David, Vicini, Zeltner, and
  Jakob]{nimier2019mitsuba}
Merlin Nimier-David, Delio Vicini, Tizian Zeltner, and Wenzel Jakob.
\newblock Mitsuba 2: A retargetable forward and inverse renderer.
\newblock \emph{ACM Transactions on Graphics (TOG)}, 38\penalty0 (6):\penalty0
  1--17, 2019.

\bibitem[Oberweger et~al.(2015)Oberweger, Wohlhart, and
  Lepetit]{oberweger2015training}
Markus Oberweger, Paul Wohlhart, and Vincent Lepetit.
\newblock Training a feedback loop for hand pose estimation.
\newblock In \emph{Proceedings of the IEEE international conference on computer
  vision}, pages 3316--3324, 2015.

\bibitem[Oikonomidis et~al.(2011{\natexlab{a}})Oikonomidis, Kyriazis, and
  Argyros]{oikonomidis2011efficient}
Iason Oikonomidis, Nikolaos Kyriazis, and Antonis~A Argyros.
\newblock Efficient model-based 3d tracking of hand articulations using kinect.
\newblock In \emph{BMVC}, volume~1, page~3, 2011{\natexlab{a}}.

\bibitem[Oikonomidis et~al.(2011{\natexlab{b}})Oikonomidis, Kyriazis, and
  Argyros]{oikonomidis2011full}
Iason Oikonomidis, Nikolaos Kyriazis, and Antonis~A Argyros.
\newblock Full dof tracking of a hand interacting with an object by modeling
  occlusions and physical constraints.
\newblock In \emph{2011 International Conference on Computer Vision}, pages
  2088--2095. IEEE, 2011{\natexlab{b}}.

\bibitem[Oikonomidis et~al.(2010)Oikonomidis, Kyriazis, and
  Argyros]{oikonomidis2010markerless}
Iasonas Oikonomidis, Nikolaos Kyriazis, and Antonis~A Argyros.
\newblock Markerless and efficient 26-dof hand pose recovery.
\newblock In \emph{Asian Conference on Computer Vision}, pages 744--757.
  Springer, 2010.

\bibitem[Panteleris and Argyros(2017)]{panteleris2017back}
Paschalis Panteleris and Antonis Argyros.
\newblock Back to rgb: 3d tracking of hands and hand-object interactions based
  on short-baseline stereo.
\newblock In \emph{Proceedings of the IEEE International Conference on Computer
  Vision Workshops}, pages 575--584, 2017.

\bibitem[Panteleris et~al.(2018)Panteleris, Oikonomidis, and
  Argyros]{panteleris2018using}
Paschalis Panteleris, Iason Oikonomidis, and Antonis Argyros.
\newblock Using a single rgb frame for real time 3d hand pose estimation in the
  wild.
\newblock In \emph{2018 IEEE Winter Conference on Applications of Computer
  Vision (WACV)}, pages 436--445. IEEE, 2018.

\bibitem[Pavlakos et~al.(2019)Pavlakos, Kolotouros, and
  Daniilidis]{pavlakos2019texturepose}
Georgios Pavlakos, Nikos Kolotouros, and Kostas Daniilidis.
\newblock Texturepose: Supervising human mesh estimation with texture
  consistency.
\newblock In \emph{Proceedings of the IEEE/CVF International Conference on
  Computer Vision}, pages 803--812, 2019.

\bibitem[Qian et~al.(2014)Qian, Sun, Wei, Tang, and Sun]{qian2014realtime}
Chen Qian, Xiao Sun, Yichen Wei, Xiaoou Tang, and Jian Sun.
\newblock Realtime and robust hand tracking from depth.
\newblock In \emph{Proceedings of the IEEE conference on computer vision and
  pattern recognition}, pages 1106--1113, 2014.

\bibitem[Qian et~al.(2020)Qian, Wang, Mueller, Bernard, Golyanik, and
  Theobalt]{HTML_eccv2020}
Neng Qian, Jiayi Wang, Franziska Mueller, Florian Bernard, Vladislav Golyanik,
  and Christian Theobalt.
\newblock {HTML: A Parametric Hand Texture Model for 3D Hand Reconstruction and
  Personalization}.
\newblock In \emph{Proceedings of the European Conference on Computer Vision
  (ECCV)}. Springer, 2020.

\bibitem[Ravi et~al.(2020)Ravi, Reizenstein, Novotny, Gordon, Lo, Johnson, and
  Gkioxari]{ravi2020pytorch3d}
Nikhila Ravi, Jeremy Reizenstein, David Novotny, Taylor Gordon, Wan-Yen Lo,
  Justin Johnson, and Georgia Gkioxari.
\newblock Accelerating 3d deep learning with pytorch3d.
\newblock \emph{arXiv:2007.08501}, 2020.

\bibitem[Rehg and Kanade(1994)]{rehg1994digiteyes}
James~M Rehg and Takeo Kanade.
\newblock Digiteyes: Vision-based hand tracking for human-computer interaction.
\newblock In \emph{Proceedings of 1994 IEEE Workshop on Motion of Non-rigid and
  Articulated Objects}, pages 16--22. IEEE, 1994.

\bibitem[Remilekun~Basaru et~al.(2017)Remilekun~Basaru, Slabaugh, Alonso, and
  Child]{remilekun2017hand}
Rilwan Remilekun~Basaru, Greg Slabaugh, Eduardo Alonso, and Chris Child.
\newblock Hand pose estimation using deep stereovision and markov-chain monte
  carlo.
\newblock In \emph{Proceedings of the IEEE International Conference on Computer
  Vision Workshops}, pages 595--603, 2017.

\bibitem[Riba et~al.(2020)Riba, Mishkin, Ponsa, Rublee, and
  Bradski]{eriba2019kornia}
Edgar Riba, Dmytro Mishkin, Daniel Ponsa, Ethan Rublee, and Gary Bradski.
\newblock Kornia: an open source differentiable computer vision library for
  pytorch.
\newblock In \emph{Winter Conference on Applications of Computer Vision}, 2020.
\newblock URL \url{https://arxiv.org/pdf/1910.02190.pdf}.

\bibitem[Romero et~al.(2009)Romero, Kjellstr{\"o}m, and
  Kragic]{romero2009monocular}
Javier Romero, Hedvig Kjellstr{\"o}m, and Danica Kragic.
\newblock Monocular real-time 3d articulated hand pose estimation.
\newblock In \emph{2009 9th IEEE-RAS International Conference on Humanoid
  Robots}, pages 87--92. IEEE, 2009.

\bibitem[Romero et~al.(2017)Romero, Tzionas, and Black]{MANO:SIGGRAPHASIA:2017}
Javier Romero, Dimitrios Tzionas, and Michael~J. Black.
\newblock Embodied hands: Modeling and capturing hands and bodies together.
\newblock \emph{ACM Transactions on Graphics, (Proc. SIGGRAPH Asia)}, 2017.

\bibitem[Rudnev et~al.(2020)Rudnev, Golyanik, Wang, Seidel, Mueller, Elgharib,
  and Theobalt]{rudnev2020eventhands}
Viktor Rudnev, Vladislav Golyanik, Jiayi Wang, Hans-Peter Seidel, Franziska
  Mueller, Mohamed Elgharib, and Christian Theobalt.
\newblock Eventhands: Real-time neural 3d hand reconstruction from an event
  stream.
\newblock \emph{arXiv preprint arXiv:2012.06475}, 2020.

\bibitem[Simonyan and Zisserman(2014)]{simonyan2014very}
Karen Simonyan and Andrew Zisserman.
\newblock Very deep convolutional networks for large-scale image recognition.
\newblock \emph{arXiv preprint arXiv:1409.1556}, 2014.

\bibitem[Smith et~al.(2020)Smith, Wu, Wen, Peluse, Sheikh, Hodgins, and
  Shiratori]{smith2020constraining}
Breannan Smith, Chenglei Wu, He~Wen, Patrick Peluse, Yaser Sheikh, Jessica~K
  Hodgins, and Takaaki Shiratori.
\newblock Constraining dense hand surface tracking with elasticity.
\newblock \emph{ACM Transactions on Graphics (TOG)}, 2020.

\bibitem[Spurr et~al.(2021{\natexlab{a}})Spurr, Dahiya, Zhang, Wang, and
  Hilliges]{spurr2021self}
Adrian Spurr, Aneesh Dahiya, Xucong Zhang, Xi~Wang, and Otmar Hilliges.
\newblock Self-supervised 3d hand pose estimation from monocular rgb via
  contrastive learning.
\newblock \emph{arXiv preprint arXiv:2106.05953}, 2021{\natexlab{a}}.

\bibitem[Spurr et~al.(2021{\natexlab{b}})Spurr, Molchanov, Iqbal, Kautz, and
  Hilliges]{spurr2021adversarial}
Adrian Spurr, Pavlo Molchanov, Umar Iqbal, Jan Kautz, and Otmar Hilliges.
\newblock Adversarial motion modelling helps semi-supervised hand pose
  estimation.
\newblock \emph{arXiv preprint arXiv:2106.05954}, 2021{\natexlab{b}}.

\bibitem[Sridhar et~al.(2013)Sridhar, Oulasvirta, and
  Theobalt]{sridhar2013interactive}
Srinath Sridhar, Antti Oulasvirta, and Christian Theobalt.
\newblock Interactive markerless articulated hand motion tracking using rgb and
  depth data.
\newblock In \emph{Proceedings of the IEEE international conference on computer
  vision}, pages 2456--2463, 2013.

\bibitem[Tagliasacchi et~al.(2015)Tagliasacchi, Schr{\"o}der, Tkach, Bouaziz,
  Botsch, and Pauly]{tagliasacchi2015robust}
Andrea Tagliasacchi, Matthias Schr{\"o}der, Anastasia Tkach, Sofien Bouaziz,
  Mario Botsch, and Mark Pauly.
\newblock Robust articulated-icp for real-time hand tracking.
\newblock In \emph{Computer Graphics Forum}, volume~34, pages 101--114. Wiley
  Online Library, 2015.

\bibitem[Tang et~al.(2014)Tang, Jin~Chang, Tejani, and Kim]{tang2014latent}
Danhang Tang, Hyung Jin~Chang, Alykhan Tejani, and Tae-Kyun Kim.
\newblock Latent regression forest: Structured estimation of 3d articulated
  hand posture.
\newblock In \emph{Proceedings of the IEEE conference on computer vision and
  pattern recognition}, pages 3786--3793, 2014.

\bibitem[Thompson and Galata(2020)]{thompson2020hand}
Peter Thompson and Aphrodite Galata.
\newblock Hand tracking from monocular rgb with dense semantic labels.
\newblock In \emph{2020 15th IEEE International Conference on Automatic Face
  and Gesture Recognition (FG 2020)}, pages 394--401. IEEE, 2020.

\bibitem[Tkach et~al.(2017)Tkach, Tagliasacchi, Remelli, Pauly, and
  Fitzgibbon]{tkach2017online}
Anastasia Tkach, Andrea Tagliasacchi, Edoardo Remelli, Mark Pauly, and Andrew
  Fitzgibbon.
\newblock Online generative model personalization for hand tracking.
\newblock \emph{ACM Transactions on Graphics (ToG)}, 36\penalty0 (6):\penalty0
  1--11, 2017.

\bibitem[Tzionas et~al.(2016)Tzionas, Ballan, Srikantha, Aponte, Pollefeys, and
  Gall]{tzionas2016capturing}
Dimitrios Tzionas, Luca Ballan, Abhilash Srikantha, Pablo Aponte, Marc
  Pollefeys, and Juergen Gall.
\newblock Capturing hands in action using discriminative salient points and
  physics simulation.
\newblock \emph{International Journal of Computer Vision}, 118\penalty0
  (2):\penalty0 172--193, 2016.

\bibitem[W{\"o}hlke et~al.(2018)W{\"o}hlke, Li, and Lee]{wohlke2018model}
Jan W{\"o}hlke, Shile Li, and Dongheui Lee.
\newblock Model-based hand pose estimation for generalized hand shape with
  appearance normalization.
\newblock \emph{arXiv preprint arXiv:1807.00898}, 2018.

\bibitem[Zimmermann and Brox(2017)]{zb2017hand}
Christian Zimmermann and Thomas Brox.
\newblock Learning to estimate 3d hand pose from single rgb images.
\newblock Technical report, arXiv:1705.01389, 2017.
\newblock URL \url{https://lmb.informatik.uni-freiburg.de/projects/hand3d/}.
\newblock https://arxiv.org/abs/1705.01389.

\bibitem[Zimmermann et~al.(2021)Zimmermann, Argus, and
  Brox]{zimmermann2021contrastive}
Christian Zimmermann, Max Argus, and Thomas Brox.
\newblock Contrastive representation learning for hand shape estimation.
\newblock \emph{arXiv preprint arXiv:2106.04324}, 2021.

\end{thebibliography}
\end{document}